# Testing AI on language comprehension tasks reveals insensitivity to underlying meaning

Vittoria Dentella[1], Fritz Günther[2], Elliot Murphy[3], Gary Marcus[4], Evelina Leivada[5,6]

[1] Departament d'Estudis Anglesos i Alemanys, Universitat Rovira i Virgili, Tarragona, Spain
[2] Institut für Psychologie, Humboldt-Universitat zu Berlin, Berlin, Germany
[3] Department of Neurosurgery, UTHealth, Houston, Texas, USA
[4] Department of Psychology, New York University, New York, USA
[5] Departament de Filologia Catalana, Universitat Autònoma de Barcelona, Barcelona, Spain
[6] Institució Catalana de Recerca i Estudis Avançats (ICREA), Barcelona, Spain

## Abstract

Large Language Models (LLMs) are recruited in applications that span from clinical assistance and legal support to question answering and education. Their success in specialized tasks has led to the claim that they possess human-like linguistic capabilities related to compositional understanding and reasoning. Yet, reverse-engineering is bound by Moravec's Paradox, according to which easy skills are hard. We systematically assess 7 state-of-the-art models on a novel benchmark. Models answered a series of comprehension questions, each prompted multiple times in two settings, permitting one-word or open-length replies. Each question targets a short text featuring high-frequency linguistic constructions. To establish a baseline for achieving human-like performance, we tested 400 humans on the same prompts. Based on a dataset of n=26,680 datapoints, we discovered that LLMs perform at chance accuracy and waver considerably in their answers. Quantitatively, the tested models are outperformed by humans, and qualitatively their answers showcase distinctly non-human errors in language understanding. We interpret this evidence as suggesting that, despite their usefulness in various tasks, current AI models fall short of understanding language in a way that matches humans, and we argue that this may be due to their lack of a compositional operator for regulating grammatical and semantic information.

## Introduction

Large Language Models (LLMs) are neural networks trained on generating probability distributions over natural language data. Through interfaces that allow direct interaction with users, LLMs (i.e., the underlying models together with their respective interface setup) perform tasks that span from translation to answering a wide range of general queries (Bubeck et al., 2024), spanning domains as diverse as law (Nay et al., 2024), medicine (Singhal et al., 2023;

Sandmann et al., 2024), and chemistry (Jablonka et al., 2024). Yet, good performance in tasks that require memorizing specialized knowledge is not necessarily grounded in a solid understanding of language, such that LLMs may fail at comparatively easier tasks (*Moravec's paradox*; Moravec, 1988, 1999; Pinker, 1994). Reverse-engineering is harder for simple, effortless tasks that our minds do best (Minsky, 1986); and understanding language is easy for humans, with even eighteen-month-olds demonstrating an understanding of complex grammatical relations (Perkins & Lidz, 2021). Our species is characterized by an irrepressible predisposition to acquire language, to seek meaning beneath the surface of words, and to impose on linear sequences a surprisingly rich array of hierarchical structure and relations (Dehaene et al., 2022).

Based on their success in various applications, LLMs have been linked with human-like capabilities such as advanced reasoning (OpenAI on ChatGTP and GPT-4; OpenAI, 2024), understanding across modalities (Google on Gemini; Pichai & Hassabis, 2023), and common sense (Bubeck et al., 2024). Some scholars have claimed that LLMs approximate human cognition, understand language to the point that their performance aligns with or even surpasses that of humans, and are good models of human language (Piantadosi & Hill, 2022; Mahowald et al., 2024; Hu et al., in press). However, much evidence exists pointing to possible inconsistencies in model performance. Models can both produce highly fluent, semantically coherent outputs (Radford et al., 2019; Brown et al., 2020; Bojic et al., 2024; Bubeck et al., 2024) whilst also struggling with fundamental syntactic or semantic properties of natural language (Bender & Koller, 2020; Ettinger, 2020; Rassin et al., 2022; Dentella et al., 2023; Leivada et al., 2023; Moro et al., 2023; Murphy et al., 2024). Observing errors in the linguistic performance of LLMs begs the question of how their application in various tasks, such as replying to a medical or legal question, is affected by possible language limitations, and may in fact be achieved through some series of computational steps entirely remote from a cognitively plausible architecture for human language (Collins, 2024).

While shortcomings in LLM performance do not negate their usefulness as tools for specific applied uses, they do invite serious concerns over what guides output generation: is it (i) context-sensitive text parsing (i.e., the ability to pair specific linguistic forms with their corresponding meanings) in a way that generalizes across contexts (Marcus & Davis, 2019; Sinha et al., 2023), or (ii) the mechanistic leveraging of artifacts present in the training data (Kandpal et al., 2022), hence giving merely the illusion of competence? At present, the dominant means of evaluating LLMs consists in employing their (well-formed) outputs as a basis to infer human-like language skills (*modus ponens*, for Guest & Martin, 2023). For instance, accurate performance in language-oriented tasks and benchmarks (e.g., Dua et al., 2019; Wang et al., 2019; Srivastava et al., 2023) is used to conclude that the LLMs have succeeded not only at the specific task they performed, but also at learning the general knowledge necessary in order to carry out that task (cf. Yogatama et al., 2019). This line of reasoning lies at the basis for holding LLMs as cognitive theories (Piantadosi, 2023; Mahowald et al., 2024).

On the other hand, assuming competence for LLMs over all morphosyntactic, semantic and pragmatic processes involved in language comprehension clashes with their inability to consistently deploy the knowledge that is attributed to them (Dentella et al., 2023; Leivada et al., 2023; Marcus & Davis, 2023). We list the grammaticality judgments (i.e., judgments about whether a prompt complies with or deviates from the language patterns that the model has internalized) of two frontier LLMs on linguistic prompts that are less frequent in day-to-day language and, consequently, are considerably less likely to form part of the training data (Figure 1). While in humans the influence of cognitive factors (e.g., working memory limitations or distraction) can affect language processing in ways that may result in non-target grammaticality judgments, humans are able to correctly process the stimuli upon reflection; that is, we can engage in 'deep' processing after an initial 'shallow' parse (Karimi & Ferreira, 2016). For LLMs, however, there is no analogous straightforward explanation of their systematic language errors.

| Phenomenon | Prompt | GPT-3 | ChatGPT-3.5 | Linguistic analysis |
|---|---|---|---|---|
| Identity Avoidance violation | Dogs dogs dog dog dogs | Inaccurate | Inaccurate | Both prompts are grammatical and have the same structure: in English, *dog* can be either a noun (a dog) or a verb (to dog). The first prompt can be paraphrased as *Dogs that are followed by dogs follow themselves other dogs.* |
| | Fish fishermen catch eat worms | Inaccurate | Inaccurate | |
| Agreement attraction | *The key to the drawers are on the table | Accurate | Accurate | Both prompts are ungrammatical and have the same structure: the singular subject (*the key* and *the village*, respectively) does not agree with the plural verb (*are* and *have met*). |
| | *The village of the fundamentally flawed ideas have not met the criteria | No data. Model deprecated | Inaccurate | |
| Comparative sentences | *More people have been to Russia than I have | Inaccurate | Inaccurate | Both prompts are ungrammatical: the subject of the main clause (*more people* and *fewer people*, respectively) require a cardinal set for comparison purposes, whereas the subject of the embedded clause, i.e., *I*, is singular. |
| | *Fewer people have been to Russia than I have | Inaccurate | Inaccurate | |
| Missing verb | The patient the nurse the clinic had hired admitted met Jack | Inaccurate | Inaccurate | The second prompt is ungrammatical, as one verb phrase is missing. Under the processing demands induced by center embedding, this violation may go unnoticed. |
| | *The doctor the nurse the clinic had hired met Jack | Inaccurate | Inaccurate | |
| Semantic anomalies | When an airplane crashes at the border of two countries and debris is scattered on the ground of both, where should the survivors be buried? | Inaccurate | Inaccurate | Both prompts are semantically anomalous: survivors are not buried. In the first prompt, the anomaly is induced at the sentence level; in the second prompt, the violation is rendered through a locally anomalous noun phrase (i.e., *surviving dead*). |
| | If a plane crashed on the border of US and Canada, where should we bury the surviving dead? | Inaccurate | Inaccurate | |

Figure 1: GPT-3 (text-davinci-002) and ChatGPT-3.5 performance on a grammaticality judgment task that involves low-frequency constructions. Inaccurate responses are marked in red. Accurate responses are marked in green. Full model replies and detailed linguistic analyses are available (osf.io/dfgmr/?view_only=063bcf2314c547199063e87da79c32b3).

While it is incontestable that LLMs offer a potentially useful interactive tool that mimics human conversation, it remains to be established whether their ability to understand language is on a par with that of humans. To answer, we investigate the ability of 7 state-of-the-art LLMs in a comprehension task that features prompts whose linguistic complexity is purposely kept at a minimum, e.g. (1).

(1) John deceived Mary and Lucy was deceived by Mary. In this context, did Mary deceive Lucy?

The benefit of this methodology is that it tests LLMs in the very task for which they have been trained: respond to questions based on the input. While previous research testing LLMs on grammaticality judgments has found that their performance is not comparable to that of humans (Dentella et al., 2023), the results of this reported method are not unequivocally convincing. Some have argued that judgment prompting is not a suitable methodology for assessing LLMs, because this is (putatively) a metalinguistic skill that LLMs do not necessarily possess (Hu & Levy, 2023; Hu et al., in press). Thus, in the present work we test LLMs in their default ability to answer questions based on input. Given their use in a range of meaning-sensitive tasks, from giving medical diagnoses to travel advice and to being “thought partners” to students (Extance, 2023; Singhal et al., 2023), it is expected that LLMs should be able to accurately parse prompts featuring high-frequency verbs and constructions.

To this end, we asked whether LLMs’ language performance is on a par with that of humans. Two research questions (RQs) emerge from this general aim: **RQ1**: Do LLMs reply to comprehension questions accurately? **RQ2**: When the same question is asked multiple times, do LLMs reply consistently?

# Results

Results are reported separately for *accuracy* (RQ1) and *stability* (RQ2). First, LLMs are compared between each other on these two variables. Then, their performance is contrasted with that of humans. (Generalized) Linear Mixed Effect Models ((G)LMMs) are used to analyze the data, using the lme4 (Bates et al., 2015) and lmerTest (Kuznetsova et al., 2017) packages in R. LLM and human datasets, together with the annotated code employed in the analyses are available (osf.io/dfgmr/?view_only=063bcf2314c547199063e87da79c32b3).

## LLMs: Accuracy

**Accuracy by model and setting** Accuracy rates by model and setting (open-length vs. one-word) are displayed in Figure 2A. The intercept in an intercept-only GLMM (accuracy ~ 1 + (1 | test_item)) is not significantly different from zero ($\beta = 0.224$, $z = 1.71$, $P = .087$), indicating at-chance performance when all LLMs are taken together. A first chi-square test between a GLMM with a main effect of model (accuracy ~ model + (1 | test_item)) and this intercept-only GLMM however finds significant differences in the performance of the 7 LLMs, justifying the introduction of a model parameter in the GLMM ($X2(6) = 202.8$, $P < .001$). The intercept of a GLMM with Bard set as baseline category is negative ($\beta = -0.688$, $z = -3.433$, $P = .001$), indicating that the performance of at least one LLM is below chance. By systematically changing the baseline category, we find that Mixtral is the first LLM to perform at chance ($\beta = 0.103$, z =

0.524, P = .600). With a mean accuracy of .66, Llama2 is the first LLM to perform above chance (β = 0.752, z = 3.720, P < .001), alongside Falcon (M = .66) and ChatGPT-4 (M = .83). Further, we find that ChatGPT-4 has a significantly higher accuracy than all other LLMs: When setting ChatGPT-4 as the baseline category, all parameters for all other models are significant (*P*s < .001) and negative (*β*s < -1.01, *z*s < -4.31).

A second chi-square test between a GLMM with a main effect of setting (accuracy ~ model + setting + (1 | test_item)) and a GLMM without it (accuracy ~ model + (1 | test_item)) finds a significant main effect of setting (X2(1) = 82.587, P < .001), with more accurate responses in the one-word setting than in the open-length setting (β = 1.037, z = 8.872 , P < .001). We additionally find an interaction between model and setting (X2(6) = 43.374, P < .001), indicating that most but not all LLMs perform more poorly in the open-length setting.

## LLMs: Stability

In order to establish the extent to which the LLMs provide stable responses, we performed an analysis at the item level (as opposed to the individual trial level for the accuracy data). Here, the responses to an item were coded as stable (= 1) if all three responses to a prompt were the same, and as unstable (= 0) if one response was different than the other two.

**Stability by model and setting** Stability rates by model and setting are displayed in Figure 2B. As was the case for accuracy, a GLMM including a parameter for model significantly outperforms a GLMM without it (X2(6) = 21.438, P = .002). In a GLMM with Bard (the LLM with the lowest stability, M = .66) as baseline category, the intercept is positive (β = 0.610, z = 2.810, P = .005), indicating that even for the most unstable model, the responses are not fully random (i.e., there is a higher likelihood of stable than unstable response patterns). Additionally, the model parameters show that stability is significantly higher for Falcon (β = 0.901, z = 2.357, P = .018; M = .83) and Gemini (β = 0.733, z = 1.976, P = .048; M = .80). Taken together with accuracy results, significantly higher stability for Falcon and Gemini respectively translates into Falcon being partially consistent in providing accurate replies, and Gemini being partially consistent in providing inaccurate replies. Employing Falcon (the most stable LLM) as baseline category for the GLMM, we find that Bard (β = -0.901, z = -2.357, P = .018) and Mixtral (β = -1.340, z = -3.559, P < .001) display less stable response patterns than Falcon, while all the other LLMs perform comparably to Falcon.

Similar to accuracy, a second chi-square test between a GLMM with a main effect of setting (stability ~ model + setting + (1 | test_item)) and a GLMM without it (stability ~ model + (1 | test_item)), finds a significant main effect of setting on stability (X2(1) = 5.160, P = .023), with answers being more stable in the one-word setting (β = 0.450, z = 2.262, P = .024). Lastly, we find no significant interaction between model and setting, indicating that the effect of setting does not differ between the different LLMs (X2(6) = 4.241, P = .644).

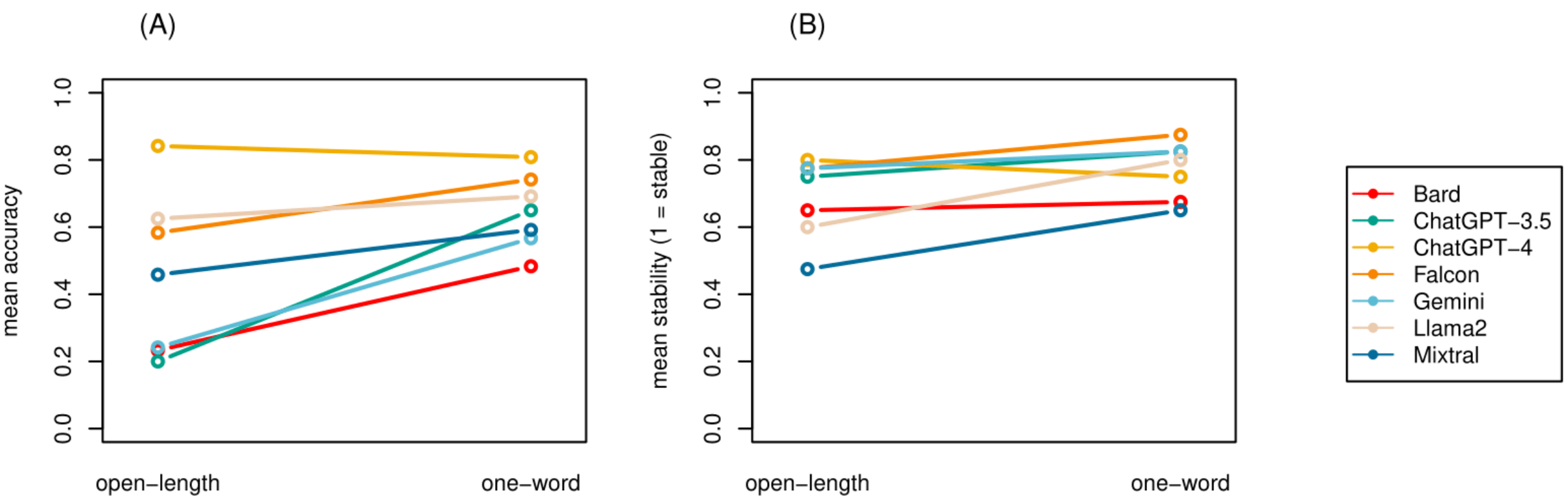

Figure 2: (A) Mean accuracy by model and setting. (B) Mean stability by model and setting.

## LLMs vs. Humans

Comparative analyses showed major differences in the performance of the two responding agents (humans vs. LLMs) in terms of both accuracy and stability (Figure 3).

**Accuracy by agent and setting** Accuracy rates by responding agent and setting are displayed in Figure 3A. A first chi-square test finds that a GLMM with a main effect of agent (accuracy ~ setting + agent + (1 | item) + (1 | unit)) significantly outperforms a GLMM without it (accuracy ~ setting + (1 | item) + (1 | unit)), justifying the introduction of an agent parameter in the GLMM ($X2(1) = 23.356$, $P < .001$). Additionally, a GLMM with an interaction between agent and setting (accuracy ~ setting * agent + (1 | item) + (1 | unit)) outperforms the GLMM that only includes them as main effects ($X2(1) = 45.265$, $P < .001$). In the GLMM including the interaction, with humans and the open-length setting as baseline category ($\beta = 3.270$, $z = 16.190$, $P < .001$ for the intercept), we find that LLMs perform significantly worse than humans in the open-length setting ($\beta = -3.508$, $z = -6.103$, $P < .001$), that humans do not perform better in the one-word setting as opposed to the open-length setting ($\beta = -0.228$, $z = -1.561$, $P = .118$), and that the gap between humans and LLMs is significantly smaller for responses in the one-word setting, indicating a discrepancy between the response settings for LLMs that is not observed for humans ($\beta = 1.276$, $z = 6.833$, $P < .001$; see Figure 3).

**Stability by agent and setting** Stability rates by responding agent and setting are displayed in Figure 3B. As was the case for accuracy, a first chi-square test finds that a GLMM with a main effect of agent (stability ~ setting + agent + (1 | item) + (1 | unit)) outperforms a GLMM without it ($X2(1) = 10.515$, $P = .001$). A second chi-square test additionally finds that a GLMM with an interaction between agent and setting (stability ~ setting * agent + (1 | item) + (1 | unit)) significantly outperforms a GLMM which includes agent and setting as main effects ($X2(1) = 4.903$, $P = .027$). In the GLMM including the interaction with humans and the open-length setting again as baseline category (β = 2.286, z = 17.368, P < .001 for the intercept), we find that LLMs are less stable than humans in the open-length setting (β = -1.385, z = -3.800, P < .001), that humans are not less stable in the open-length setting compared to the one-word setting (β = -0.038, z = -0.362, P = .717), and that in the one-word setting, the stability gap of LLMs with humans is smaller than in the open-length setting (again indicating a discrepancy between the settings for LLMs that is not observed for humans; β = 0.507, z = 2.234, P = 0.026; see Figure 3).

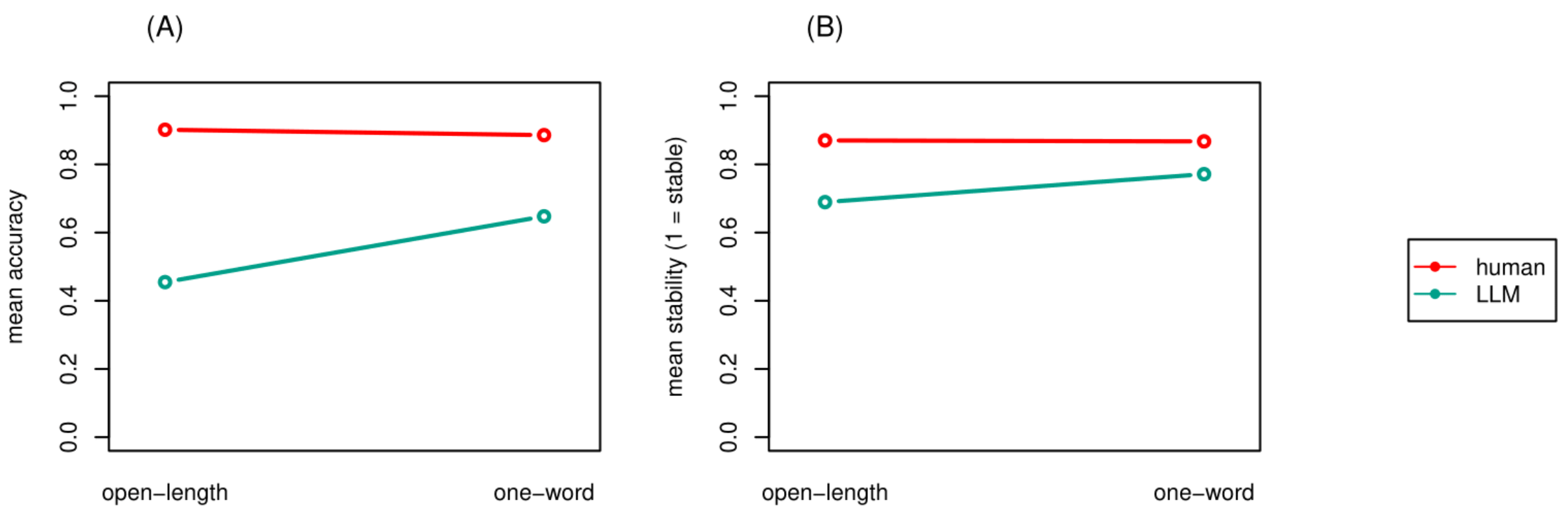

Figure 3: (A) Mean accuracy by responding agent and setting. (B) Mean stability by responding agent and setting.

Lastly, anticipating possible arguments that LLMs as a category should be represented by their best performing representative, ChatGPT-4 (the best performing LLM among those tested) can be compared to the best performing humans (as opposed to the averaged performance of all humans). Given that n = 51 human subjects perform at ceiling in terms of accuracy and therefore stability, while ChatGPT-4 does not perform at ceiling on either variable, it follows that even the best performing LLM performs significantly worse than the best performing humans.

## Discussion

This work sheds light on what guides output generation in LLMs: (i) the human-like ability to parse and understand written text, or (ii) the leveraging of artifacts in the training data. To do so, we test seven state-of-the-art LLMs on comprehension questions targeting sentences that feature high-frequency constructions and lexical items, and whose linguistic complexity is kept at a minimum. Particularly, we aim at determining whether LLMs generate answers that are both accurate (RQ1) and stable over repeated trials (RQ2). Systematic testing shows that the average performance of LLMs as a group is at chance for accuracy, and that LLMs are relatively unstable in their answers. On the other hand, humans tested on the same comprehension questions provide mostly accurate answers (RQ1) which almost never change when a question is repeatedly prompted (RQ2). Importantly, these differences between LLMs and humans are pronounced even though the models' replies are favorably scored, ignoring any possible inconsistencies they contain (i.e., in the one-word setting; see section 'Methods').

Language parsing, referring to the ability to comprehend and produce language by assigning strings of symbols with meaning, is a distinctively human ability (Hauser et al., 2002). This explains why humans in our experiment responded accurately and without alternating answers when a question is asked multiple times, or with different instructions. For humans, failure to reach at-ceiling performance is explained by the interplay of processing factors (Karimi & Ferreira, 2016) and interspeaker variation which affects interpretative choices. For instance, given the prompt *Steve hugged Molly and Donna was kissed by Molly. In this context, was Molly kissed?* the answer is "no", though some might assign the verb *to kiss* a reciprocal reading consequently replying "yes". Nonetheless, despite processing limitations and the variability induced by subjective choices, humans cohesively outperform LLMs both in terms of accuracy and stability, thus confirming that testing LLMs on foundational tasks can reveal shortcomings which remain concealed in more sophisticated settings (Pinker, 1994).

The LLM outputs differ from human answers both quantitatively and qualitatively. Quantitatively, the averaged accuracy of LLMs as a group is at chance, and the models which manage to surpass the chance threshold (i.e., Falcon, Llama2, and ChatGPT-4) reach accuracy rates that are far from perfect. The best performing LLM, ChatGPT-4, performs worse than the best performing humans. Secondly, while all LLMs perform above chance in terms of stability, none of them succeeds in consistently providing the same answer to a question. Taken together, this evidence translates into an overall inability on the LLMs' part to tackle simple comprehension questions in a way that can be reasonably called 'human-like'.

Building on how their outputs show surface similarities with natural language, LLMs have been endowed with human-like language skills both in public discourse, where they are described as "understanding", "reasoning", or "intelligent", and in science, where their adoption as cognitive models, and even *theories*, is increasingly called for (Piantadosi, 2023; Mahowald et al., 2024). Our results suggest that such a framing may not be justified at present. Importantly, once a LLM has been invested with competence over a language phenomenon, without theoretically informed hypotheses at hand (Guest & Martin, 2020), the causes of its failures to perform at ceiling leave room for speculation. We thus interpret the LLMs' failure in providing accurate and stable answers in a simple comprehension task as suggesting that the models lack a real understanding of language: they produce words that are like semantic black-boxes, approximating the surface statistics of language and the more 'automatic' aspects of parsing (Mitchell & Krakauer, 2022). In addition, LLMs are not suitable for serving as theories of language since their representational capacities are virtually unbounded, rendering their representations arbitrary and not explanatorily grounded, belonging as they do to the class of Universal Functional Approximators (Yun et al., 2019) which have been shown to be capable of approximating any mathematical function (Ismailov, 2023). Consequently, LLMs bear a closer resemblance to tools such as generalized derivatives than scientific theories (see also Evanson et al., 2023).

Additionally, the differences between LLM and human performance are also qualitative in nature. First, all human participants with no exception provide one-word replies regardless of the setting. This shows alignment between subjects on the amount of information that is necessary in order to satisfactorily reply to a question (Grice, 1975), and even in this space of pragmatics humans seem to comply with principles of computational efficiency and relevance that are built out of the path that syntactic inferences carve out for them (Carston, 2022). This also marks a stark difference from LLMs which typically provide lengthy explanations, even in the one-word setting, thus defying the task instructions. In these explanations, their reasoning is broken down into logical steps which often contain irrelevant or contradicting information. For example, to the prompt "Franck read to himself and John read to himself, Anthony and Franck. In this context, was Franck read to?", Gemini replies (2):

> (2) "In this context, *it's impossible to say for sure if Franck was read to*. Here's why:
>
> 1. Limited information: While the sentence "John read to himself, Anthony and Franck" implies John read aloud to others, it doesn't clarify if Franck was one of the recipients or just present during the reading.
>
> 2. Separate actions: The sentence about Franck reading to himself is separate and doesn't necessarily connect to John's reading session.
>
> 3. Ambiguity: The phrase "read to" can have different interpretations depending on the context. It could mean reading aloud *for* someone or simply reading *in the presence of* someone.
>
> Without additional information about the specific situation, such as John's reading material, the purpose of his reading, or the relationship between the characters, it's impossible to determine if Franck was actually read to by John.
>
> Here are some possibilities based on the limited information:
>
> * Franck was read to: John read aloud to both Anthony and Franck, and Franck listened.
>
> * Franck wasn't read to: John read aloud only to Anthony, and Franck was simply present during the reading.
>
> * Franck listened to himself: Both sentences describe independent reading activities, with Franck reading silently to himself and John reading aloud to others.
>
> Therefore, the answer remains *uncertain* due to the lack of sufficient context."

The correct answer to the question is “yes”. Gemini not only gives the wrong answer, but it also answers redundantly and mixes unrelated matters about the nature of the characters and the type of reading material, none of which plays a role in answering the question.

The problem of redundant, semantically irrelevant information is not a trivial one. Our testing showed that obtaining one-word replies from the tested LLMs was possible only in a reduced number of cases (mostly in ChatGPT-4 and Falcon). While this issue was ignored for the purposes of coding in the one-word setting, because we wanted to give the models the benefit of the doubt, it further showcases how LLMs’ outputs are sensitive to the meaning of prompts only in a superficial manner (Du et al., 2024). While pre-trained LLMs might not have the skills necessary for following specific instructions such as the one we provided (i.e., reply using only one word), it should be noted that ChatGPT-3.5, ChatGPT-4, Bard and Gemini include RLHF in their training, which means that their performance in this experiment benefitted from previous targeted interventions aimed at fixing incongruent outputs and maximizing compliance to user requests; yet, these models’ replies fail to conform to task demands.

If the LLMs’ fluent outputs are taken to entail human-like linguistic skills (i.e., from the production of a correct sentence such as “The scientist is happy”, we infer that the model has knowledge of the relevant rules that put the sentence together), then it follows that LLMs should possess mastery over all the words, constructions, and rules of use, that figure in and govern their outputs. Our results from grammaticality judgments (Figure 1) and language comprehension tasks (Figures 2-3) suggest that this is not the case: distinctly non-human errors are made. To discuss an example, when asked whether “The patient the nurse the clinic had hired admitted met Jack” is grammatically correct, GPT-3 said no, providing the wrong answer (Figure 1). This is not as surprising as is the correction the model volunteered: “The nurse had hired at the clinic a patient who admitted met Jack”. This answer contains a semantic anomaly that marks an important deviation from any answer a human would produce: a nurse hiring a patient. Even if we change the methodology from judgment prompting to simple question answering, we get similar errors that suggest that the tested models do not really understand the meaning of words. Recall that Gemini could not answer as to whether Franck has been read to (after being told that “John read to himself, Anthony and Franck”; example 2), because it lacks information “such as John's reading material”. Overall, if we attempt to map the encountered LLM errors across levels of linguistic analysis, it seems that their incidence becomes greater as we move from basic phonological form to the more complex outer layers of linguistic organization (Figure 4).

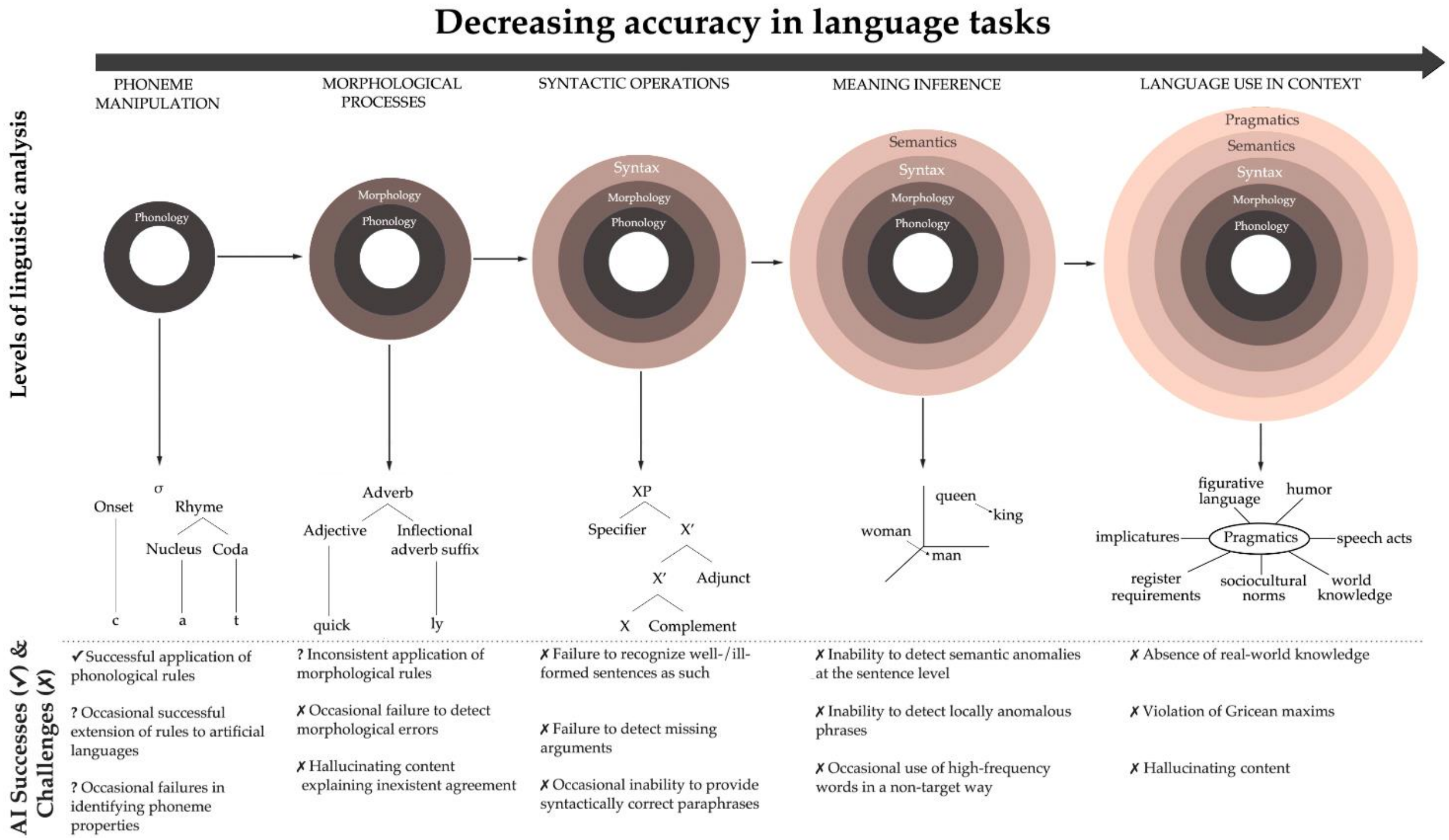

Figure 4: Mapping of the types of errors presented in Figure 1 onto levels of linguistic analysis. The observations for phonology are based on the results of Beguš et al. (2023).

While LLMs are often portrayed through wishful mnemonics that hint to human-like abilities (Mitchell, 2021), seemingly sound performance may hide flaws that are inherent to language modelling as a method: intelligence cannot in fact emerge as a side product of statistical inference (van Rooij et al., 2023), nor can the ability to understand meaning (Bender & Koller, 2020). Our results offer an argument against employing LLMs as cognitive theories (Piantadosi, 2023; Mahowald et al., 2024): the fact that LLMs are trained on natural language data does not entail that LLMs understand language; it only means that LLMs can predict some of the fossilized patterns found in training texts. Claiming a model's mastery over language because of its ability to reproduce it, is analogous to claiming that a painter *knows* a person because they can reproduce this person's face on canvas, just by looking at her pictures. Unjustly equating LLMs' abilities to those of humans in order to justify their mass-scale deployment in applied areas, in turn, results in the biases present in LLMs' training data (Birhane et al., 2023) being allowed one more display venue.

With respect to practical applications, in AI reproducing high-level reasoning is overall less demanding and needs less computational power than comparatively easier tasks (Moravec, 1999). At present, LLMs are employed in settings which require high-level reasoning. For instance, LLMs can reach high accuracy in answering medical questions (Singhal et al., 2023) and are believed to be ready for testing in real-world simulations (Mehandru et al., 2024). While LLMs in a white coat strike many as an implausible scenario, in reality LLM-based chatbots already reply to any question that any human user in real-life situations may pose, possibly including doctors. Except for warnings typically issued when ethically problematic words or politically incorrect concepts figure in prompts (Ghafouri et al., 2023), often chatbots reply without flagging any hesitations or lack of knowledge. As a result, the message conveyed is that whatever content LLMs output, it is indeed the product of true understanding and reasoning. However, as the results obtained in this work demonstrate alongside a consistent body of evidence (Rassin et al., 2022; Dentella et al., 2023; Leivada et al., 2023; Kim et al., 2024; Murphy et al., 2024), coherent, sophisticated, and refined outputs amount to patchwork in disguise.

To conclude, LLMs' fluent outputs have given rise to claims that models possess human-like language abilities, which would in turn allow them to act as cognitive theories able to provide mechanistic explanations of language processing in humans (Piantadosi, 2023; Mahowald et al., 2024). Following evidence that LLMs often produce non-human outputs (Marcus et al., 2023), in this work we tested 7 state-of-the-art LLMs on simple comprehension questions targeting short sentences, purposefully setting an extremely low bar for the evaluation of the models. Systematic testing showed that the performance of these LLMs lags behind that of humans both quantitatively and qualitatively, providing further confirmation that the easiest tasks for humans are the hardest to develop for AI (Pinker, 1994). We argue that these results invite further reflection about the standards of evaluation we adopt for claiming human-likeness in AI. Both in society and science, the role of LLMs is currently defined according to inflated human-like linguistic abilities (Leivada et al., in press), as opposed to their real capacities. If the application of internalized linguistic rules is absent and LLMs do not go beyond reproducing input patterns (Udandarao et al., 2024), all that is presented as a danger induced by the stochastic nature of LLMs can be anticipated (Bender et al., 2021). If LLMs are useful tools, understanding their limitations in a realistic way paves the way for them to be better and safer.

## Methods

The language understanding abilities of 7 LLMs are tested on a set of comprehension questions (n=40). Specifically, the employed prompts have the following characteristics: (a) sentences are affirmative (i.e., contain no negation, which is known to negatively impact LLM performance; Ettinger, 2020), (b) they are linked through coordination, (c) either 3 or 5 human entities are featured in each prompt, all identified by proper names (i.e., no pronouns), and (d) high-frequency verbs are used (e.g., to cook, to hug, to kiss), so as to maximize the chances of LLMs replying correctly. Following previous research that has suggested that LLMs fail to be stable in their answers to a question prompted repeatedly (Dentella et al., 2023), we prompted each comprehension question multiple times to assess the stability of answers. To obtain the right level of comparisons for determining human-likeness, the performance of the LLMs is compared to that of human subjects (n=400) who were tested on the same stimuli. Table 1 provides an overview of the prompts, with one sample sentence and question per condition. The full list of prompts alongside the obtained replies by both humans and LLMs is available at

https://osf.io/dfgmr/?view_only=063bcf2314c547199063e87da79c32b3.

| Sample prompt | Number of featured entities | Target answer | Number of prompts per condition |
|---|---|---|---|
| John deceived Mary and Lucy was deceived by Mary. In this context, did Mary deceive Lucy? | 3 | Yes | 10 |
| Steve hugged Molly and Donna was kissed by Molly. In this context, was Molly kissed? | 3 | No | 10 |
| Jessica and Mary were kissed by Alice. Jessica was kissed by Samuel and Andrew was kissed by Mary. In this context, was Mary kissed? | 5 | Yes | 10 |
| Bob kissed Donna and Barbara kissed Peter. Donna was hugged by Alice. In this context, was Alice hugged? | 5 | No | 10 |

Table 1: Overview of testing materials.

## LLMs

All prompts were presented to the LLMs in two settings: (i) open-length setting and (ii) one-word setting. In each setting, every prompt was repeated three times. In (i), LLMs were allowed to produce answers of any length. In (ii), the LLMs were prompted using the same materials as in (i), but the instruction "Answer using just one word" was added after every question (e.g., "John deceived Mary and Lucy was deceived by Mary. In this context, did Mary deceive Lucy? Answer using just one word"). The use of the one-word setting was intended to facilitate the LLMs, as it was noted that lengthy LLM replies often contained contradictory, irrelevant, and eventually incorrect statements. In total, each of the tested LLMs was prompted 240 times (120 times per setting). The LLMs' pool of data thus consists of n=1.680 replies (240 replies per each of the 7 LLMs).

In both settings, all replies were coded for *accuracy* (1 for accurate answers, 0 for inaccurate answers) and *stability* (1 if the three responses were either coded all as *yes* or all as *no*, 0 if at least one response was coded differently than the others). In an attempt to give the LLMs the benefit of the doubt and obtain from them the best possible result, their responses in the open-length setting which contained no clear error, but also failed to directly provide the correct answer, were coded as accurate. Additionally, due to the difficulty of obtaining one-word replies from LLMs, the decision was made to only code the first relevant piece of information provided in a reply in the one-word setting. Put another way, if a LLM correctly replied 'Yes' but this was followed by additional content that was incorrect, contradictory, or irrelevant, this was coded as accurate. From this it follows, that a less lenient coding would further amplify the differences between humans and LLMs. All prompts were merged in a unified pool, randomized and administered in random order, first in the open-length setting, and subsequently in the one-word setting.

The prompts were given to 7 LLMs, which were chosen as representative of the state of the art: ChatGPT-3.5 (Ouyang et al., 2022), ChatGPT-4 (OpenAI, 2023), Llama2 (Llama-2-70B-chat-hf; Touvron et al., 2023), Bard (Thoppilan et al., 2022), Falcon (falcon-180B-chat; Almazrouei et al., 2023), Mixtral (Mixtral-8x7B-Instruct-v0.1; Jiang et al., 2024), and Gemini (Pichai & Hassabis, 2023). ChatGPT-3.5, ChatGPT-4, Bard and Gemini include Reinforcement Learning from Human Feedback (RLHF) in their training regime. All LLMs were tested in December 2023, with the exception of Gemini which was tested in February 2024. ChatGPT-3.5 and ChatGPT-4 were prompted through OpenAI's interface; Bard and Gemini through Google's interface; and Llama2, Falcon and Mixtral through HuggingFace's interface.

### Human data

For comparison, we tested human participants (n=400; 200 F, 200 M) on the same prompts. Subjects were recruited through the crowdsourcing platform Prolific, and all participants self-identified as native speakers of English with no history of neurocognitive impairments. The experiment was carried out in accordance with the Declaration of Helsinki and was approved by the ethics committee of the Department of Psychology at Humboldt-Universität zu Berlin (application 2020-47). Participants were divided into two groups: 200 subjects (100 F, 100 M) were tested in the open-length setting, and 200 subjects (100 F, 100 M) in the one-word setting. The prompts (n=40) originally administered to the LLMs were split into two lists (each presented to half of the subjects in a condition), so that each human subject was prompted with half of the questions (n=20). Similarly to the LLMs, each participant saw each question 3 times, providing a total of n=60 replies. Overall, the human pool of data consists of n=24,000 replies. Human replies were coded for *accuracy* and *stability*. To ensure compliance, two attention checks were added that are not part of the analyses (i.e., "The door is red. In this context, is the door red?" and "The door is green. In this context, is the door blue?"). All prompts, including attention checks, were administered to each participant in random order. Data from 2 additional participants who did not provide correct answers to all attention checks were excluded from the analyses. Testing took place after participants had given informed consent to take part in the experiment. For each trial, the screen displayed the prompt (i.e., the sentence, its related question and, in the one-word setting, the additional instruction). The experiment was carried out with the jsPsych toolkit (de Leeuw et al., 2023), and had a median completion time of 13.4 minutes. Full participant data can be found at https://osf.io/dfgmr/?view_only=063bcf2314c547199063e87da79c32b3.

## References

Almazrouei, E. et al. (2023). The Falcon series of open language models. arXiv:2311.16867v2

Bates, D., Maechler, M., Bolker, B. & Walker, S. (2015). Fitting linear mixed-effects models using lme4. *Journal of Statistical Software* **67**(1), 1-48.

Beguš, G., Dąbkowski, M. & Rhodes, R. (2023). Large linguistic models: Analyzing theoretical linguistic abilities of LLMs. arXiv:2305.00948

Bender, E.M., Gebru, T., McMillan-Major, A. & Shmitchell, S. (2021). On the dangers of stochastic parrots: Can Language Models be too big? *Proceedings of the 2021 ACM Conference on Fairness, Accountability, and Transparency*, 610-623.

Bender, E.M. & Koller, A. (2020). Climbing towards NLU: On meaning, form, and understanding in the age of data. *Proceedings of the 58th Annual Meeting of the Association for Computational Linguistics*. Association for Computational Linguistics.

Birhane, A., Prabhu, V., Han, S. & Boddeti, V.N. (2023). On hate scaling laws for data-swamps. arXiv:2306.13141

Bojic, L., Kovacevic, P. & Cabarkapa, M. (2024). GPT-4 surpassing human performance in linguistic pragmatics. arXiv:2312.09545

Brown, T.B. et al. (2020). Language models are few-shot learners. arXiv:2005.14165

Bubeck, S. et al. (2024). Sparks of artificial general intelligence: Early experiments with GPT-4. arXiv:2303.12712

Carston, R. (2022). Words: Syntactic structures and pragmatic meanings. *Synthese* **200**, 430.

Collins, J. (2024). The simple reason LLMs are not scientific models (and what the alternative is for linguistics). lingbuzz/008026

Dehaene, S., Al Roumi, F., Lakretz, Y., Planton, S. & Sablé-Meyer, M. (2022). Symbols and mental programs: A hypothesis about human singularity. *Trends in Cognitive Sciences* **26**(9), 751-766.

de Leeuw, J.R., Gilbert, R.A. & Luchterhandt, B. (2023). jsPsych: Enabling an open-source collaborative ecosystem of behavioral experiments. *Journal of Open Source Softw*are **8**, 5351.

Dentella, V., Günther, F. & Leivada, E. (2023). Systematic testing of three Language Models reveals low language accuracy, absence of response stability, and a yes-response bias. *PNAS* **120**(51), e2309583120.

Du, M., He, F., Zou, N., Tao, D. & Hu, X. (2024). Shortcut learning of large language models in natural language understanding. *Communications of the ACM* **67**(1).

Dua, D., Wang, Y., Dasigi, P., Stanovsky, G., Singh, S. & Gardner, M. (2019). DROP: A reading comprehension benchmark requiring discrete reasoning over paragraphs. *Proceedings of the 2019 Conference of the North American Chapter of the Association for Computational Linguistics*, 2368-2378. Association for Computational Linguistics.

Ettinger, A. (2020). What BERT is not: Lessons from a new suite of psycholinguistic diagnostics for language models. *Transactions of the Association for Computational Linguistics* **8**, 34-48. MIT Press.

Evanson, L., Lakretz, Y. & King, J.R. (2023). Language acquisition: Do children and language models follow similar learning stages? arXiv:2306.03586

Extance, A. (2023). ChatGPT has entered the classroom: How LLMs could transform education. *Nature* **623**, 474-477.

Grice, P. (1975). Logic and conversation. In P. Cole (Ed.), *Syntax and Semantics 3: Speech Acts*. Academic Press.

Ghafouri, V., Agarwal, V., Zhang, Y., Sastry, N., Such, J. & Suarez-Tangil, G. (2023). AI in the gray: Exploring moderation policies in dialogic large language models vs. human answers in controversial topics. *Proceedings of the 32nd ACM International Conference on Information and Knowledge Management*, 556-565.

Guest, O. & Martin, A.E. (2020). How computational modeling can force theory building in psychological science. *Perspectives on Psychological Science* **16**(4), 789-802.

Guest, O. & Martin, A. (2023). On logical inference over brains, behaviour, and artificial neural networks. *Computational Brain & Behavior* **6**, 213-227.

Hauser, M.D., Chomsky, N. & Fitch, W.T. (2002). Faculty of language: What is it, who has it, and how did it evolve? *Science* **298**(5598), 1569-1579.

Hu, J. & Levy, R. (2023). Prompting is not a substitute for probability measurements in large language models. *Proceedings of the 2023 Conference on Empirical Methods in Natural Language Processing*, 5040-5060. Association for Computational Linguistics.

Hu, J., Mahowald, K., Lupyan, G., Ivanova, A. & Levy, R. (in press). Language models align with human judgments on key grammatical constructions. *PNAS*.

Ismailov, V.E. (2023). A three layer neural network can represent any multivariate function. *Journal of Mathematical Analysis and Applications* **523**(1), 127096.

Jablonka, K.M, Schwaller, P., Ortega-Guerrero, A. & Smit, B. (2024). Leveraging large language models for predictive chemistry. *Nature Machine Intelligence* **6**, 161-169.

Jiang, A.Q. et al. (2024). Mixtral of experts. arXiv:2401.04088v1

Kandpal, N., Deng, H., Roberts, A., Wallace, E., & Raffel, C. (2023). Large Language Models struggle to learn long-tail knowledge. *Proceedings of the 40th International Conference on Machine Learning*.

Karimi, H. & Ferreira, F. (2016). Good-enough linguistic representations and online cognitive equilibrium in language processing. *The Quarterly Journal of Experimental Psychology* **69**, 1013-1040.

Kim, Y. et al. (2024). Fables: Evaluating faithfulness and content selection in book-length summarization. arXiv:2404.01261v1

Kuznetsova, A., Brockhoff, P.B. & Christensen, R.H.B. (2017). lmerTest package: Tests in linear mixed effects models. *Journal of Statistical Software* **82**(13), 1-26.

Leivada, E., Günther, F. & Dentella, V. (in press). Reply to Hu et al: Applying different evaluation standards to humans vs. Large Language Models overestimates AI performance. *PNAS*.

Leivada, E., Murphy, E. & Marcus, G. (2023). DALL-E 2 fails to reliably capture common syntactic processes. *Social Sciences & Humanities Open* **8**, 100648.

Mahowald, K., Ivanova, A.I., Blank, I.A., Kanwisher, N., Tenenbaum, J.B. & Fedorenko, E. (2024). Dissociating language and thought in large language models: A cognitive perspective. *Trends in Cognitive Sciences* **19**, S1364-6613(24)00027-5.

Marcus, G. & Davis, E. (2019). *Rebooting AI: Building Artificial Intelligence We Can Trust*. Pantheon Books.

Marcus, G. & Davis, E. (2023). Hello, multimodal hallucinations. Substack. https://garymarcus.substack.com/p/hello-multimodal-hallucinations

Marcus, G., Leivada, E. & Murphy, E. (2023). A sentence is worth a thousand pictures: Can large language models understand human language? arXiv:2308.00109

Mehandru, N., Miao, B.Y., Almaraz, E.R. et al. (2024). Evaluating large language models as agents in the clinic. *Npj Digital Medicine* **7**, 84.

Minsky, M. (1986). *The Society of Mind*. Simon and Schuster.

Mitchell, M. (2021). Why AI is harder than we think. *Proceedings of the Genetic and Evolutionary Computation Conference*.

Mitchell, M. & Krakauer, D.C. (2022). The debate over understanding in AI's large language models. *PNAS* **120**(13), e2215907120.

Moravec, H. (1988). *Mind Children*. Harvard University Press.

Moravec, H. (1999). Rise of the robots. *Scientific American* **281**(6), 124-135.

Moro, A., Greco, M. & Cappa, S. (2023). Large languages, impossible languages and human brains. *Cortex* **167**, 82-85.

Murphy, E., de Villiers, J. & Lucero Morales, S. (2024). A comparative investigation of compositional syntax and semantics in DALL-E 2. arXiv:2403.12294

Nay, J.J. et al. (2024). Large language models as tax attorneys: A case study in legal capabilities emergence. *Philosophical Transactions of the Royal Society A* **382**, 20230159.

OpenAI (2023). GPT-4 technical report. arXiv:2303.08774

OpenAI (2024; April 30). https://openai.com/gpt-4

Ouyang, L. et al. (2022). Training language models to follow instructions with human feedback. arXiv:2203.02155

Perkins, L. & Lidz, J. (2021). Eighteen-month-old infants represent nonlocal syntactic dependencies. *PNAS* **118**(41), e2026469118.

Piantadosi, S. (2023). Modern language models refute Chomsky's approach to language. lingbuzz/007180

Piantadosi, S.T. & Hill, F. (2022). Meaning without reference in large language models. *Proceedings of the 36th Conference on Neural Information Processing Systems*.

Pichai, S. & Hassabis, D. (2023). Introducing Gemini: Our largest and most capable AI model. Google blog. https://blog.google/technology/ai/google-gemini-ai/#sundar-note

Pinker, S. (1994). *The Language Instinct*. William Morrow and Company.

Radford, A., Wu, J., Child, R., Luan, D., Amodei, D. & Sutskever, I. (2019). Language models are unsupervised multitask learners. https://d4mucfpksywv.cloudfront.net/better-language-models/language_models_are_unsupervised_multitask_learners.pdf

Rassin, R., Ravfogel, S. & Goldberg, Y. (2022). DALLE-2 is seeing double: Flaws in word-to-concept mapping in Text2Image models. *Proceedings of the Fifth BlackboxNLP: Analyzing and Interpreting Neural Networks for NLP*, 335-345. Association for Computational Linguistics.

Sandmann, S., Riepenhausen, S., Plagwitz, L. & Varghese, J. (2024). Systematic analysis of ChatGPT, Google search and Llama 2 for clinical decision support tasks. *Nature Communications* **15**, 2050.

Singhal, K. et al. (2023). Large language models encode clinical knowledge. *Nature* **620**, 172-180.

Sinha, K., Gauthier, J., Mueller, A., Misra, K., Fuentes, K., Levy, R. & Williams, A. (2023). Language model acceptability judgements are not always robust to context. *Proceedings of the 61st Annual Meeting of the Association for Computational Linguistics* **1**, 6043-6063. Association for Computational Linguistics.

Srivastava, A. et al. (2023). Beyond the imitation game: Quantifying and extrapolating the capabilities of language models. *Transactions on Machine Learning Research*, 1-95.

Thoppilan, R. et al. (2022). LaMDA: Language models for dialog applications. arXiv:2201.08239

Touvron, H. et al. (2023). LLaMA: Open and efficient foundation language models. arXiv:2302.13971

Udandarao, V. et al. (2024). No “zero-shot” without exponential data: Pretraining concept frequency determines multimodal model performance. arXiv:2404.04125v1

van Rooij, I., Guest, O., Adolfi, F., de Haan, R., Kolokolova, A. & Rich, P. (2023). Reclaiming AI as a theoretical tool for cognitive science. psyarXiv:10.31234/osf.io/4cbuv

Wang, A., Pruksachatkun, Y., Nangia, N., Singh, A., Michael, J., Hill, F., Levy, O. & Bowman, S. (2019). SuperGLUE: A stickier benchmark for general-purpose language understanding systems. In H. Wallach, H. Larochelle, A. Beygelzimer, F. d’ Alché Buc, E. Fox & R. Garnett (Eds.), *Advances in Neural Information Processing Systems* **32**, 3266-3280. Curran Associates Inc.

Yogatama, D. et al. (2019). Learning and evaluating general linguistic intelligence. arXiv:1901.11373

Yun, C., Bhojanapalli, S., Rawat, A.S., Reddi, S.J. & Kumar, S. (2019). Are transformers universal approximators of sequence-to-sequence functions? arXiv:1912.10077